\documentclass{article}

\PassOptionsToPackage{numbers, sort&compress}{natbib}

 \usepackage[main, final]{neurips_2026}
 \makeatletter
\renewcommand{\@noticestring}{9th Workshop on Machine Learning and the Physical Sciences (ML4PS 2026).}
\makeatother
\usepackage[utf8]{inputenc} % allow utf-8 input
\usepackage[T1]{fontenc}    % use 8-bit T1 fonts
\usepackage{hyperref}       % hyperlinks
\usepackage{url}            % simple URL typesetting
\usepackage{booktabs}       % professional-quality tables
\usepackage{amsfonts}       % blackboard math symbols
\usepackage{nicefrac}       % compact symbols for 1/2, etc.
\usepackage{microtype}      % microtypography
\usepackage{xcolor}         % colors
\usepackage{amsmath}
\usepackage{booktabs}
\usepackage{siunitx}
\usepackage{graphicx}

\title{A Conditional Normalizing Flow Prior for Auroral Electron Energy Spectra}

\author{
Nazar Misyats$^{1}$ \qquad Mizuki Fukizawa$^{2}$ \qquad Yoshimasa Tanaka$^{2}$ \qquad Satoshi Ikehata$^{3}$\\
$^{1}$University of British Columbia, Canada
\qquad $^{2}$National Institute of Polar Research, Japan\\
$^{3}$National Institute of Informatics, Japan
\\
{\tt\small nmisyats@ece.ubc.ca, \{fukizawa.mizuki, ytanaka\}@nipr.ac.jp, sikehata@nii.ac.jp}
}

\begin{document}

\maketitle

\begin{abstract}

Reconstructing the energy distribution of precipitating auroral electrons from ground-based
optical and radar measurements is an ill-posed inverse problem. Existing methods return a
single deterministic spectrum per location and control the ambiguity with a hand-designed
regularizer. We show that the spectral smoothness penalty used by recent neural-field
reconstructions is a quadratic form in the log-flux vector, and therefore an implicit
zero-mean Gaussian prior that constrains spectral curvature alone, leaving the overall flux
level and the spectral slope unconstrained. We propose to replace it with a learned prior:
a conditional normalizing flow over 19-channel electron spectra, trained on 216
satellite-days of calibrated DMSP SSJ measurements and conditioned on magnetic latitude,
magnetic local time and the planetary $K_p$ index. On held-out satellite-days the flow
assigns measured spectra with signal in every channel 18 times the density of the best
Gaussian baseline and 1400 times that of the smoothness penalty. Samples drawn at
held-out contexts reproduce the observed flux distribution and the latitude structure of
the auroral oval.

\end{abstract}

\section{Introduction}

Auroras are light emissions caused by energetic electrons precipitating along Earth's magnetic field lines and interacting with the constituents of the upper atmosphere. The energy distribution of these precipitating electrons, called the precipitating electron flux, relates directly to the physical cause of their precipitation, and provides important insights on magnetosphere-ionosphere coupling and larger scale geophysical processes.

Satellite campaigns or sounding rockets provide direct in-situ measurements of the electron population in the ionosphere.
However, these instruments only capture the electron population along their trajectory, yielding spatially and temporally localized measurements. On the other hand, optical observations of auroras from ground-based camera networks such as ALIS \citep{brandstromAuroralLargeImaging2003}, MIRACLE \citep{sangalliPerformanceStudyNew2011} and WMI \citep{ogawaDevelopmentLowcostMultiwavelength2020} in Scandinavia, and THEMIS GBO \citep{mendeTHEMISArrayGroundbased2008} in Canada, record the continuous spatial and temporal evolutions of the auroral emissions caused by these electrons at a much higher rate and resolution than in-situ techniques.

Aurora computed tomography (ACT) performs a tomographic inversion of multi-view observations of auroras in order to recover the 3D auroral emission field \citep{freyAuroralEmissionProfiles1996, freyOpticalTomographyAurora1998, asoAnalysisAuroraStereo1990, nygrenApplicationStochasticInversion1996, enellTomographylikeRetrievalAuroral2012}.
The spatially varying energy distribution of precipitating electrons producing these emissions can then be obtained by inverting a physical emission model, either as a separate step or jointly with the tomographic reconstruction \citep{janhunenReconstructionElectronPrecipitation2001, partamiesTestingInversionMethod2004, hirschReconstructionFineScaleAuroral2016, fukizawaReconstructionPrecipitatingElectrons2022}.
A recent work \citep{misyatsFastScalableReconstruction2026} instead represents the precipitating flux of electrons at the top of the atmosphere with a neural field which maps horizontal positions to local energy histograms and is optimized via a differentiable atmospheric and image formation model.

These techniques however produce deterministic energy distributions, while the inversion is severely ill-posed. Many spectra can produce the same observations, and a single reconstruction gives no measure of which spectral features the data actually constrain. Current techniques control this ambiguity with a manually tuned regularizer, which does not guarantee a physically plausible reconstruction. In particular, the spectral smoothness penalty of \cite{misyatsFastScalableReconstruction2026} constrains the curvature of the spectra only. In this work, we propose a conditional normalizing flow over electron energy spectra, trained on DMSP SSJ satellite data and conditioned on geomagnetic parameters, which could replace this penalty as a prior in auroral tomography. We show that it accurately models the distribution of spectra on held-out days, and leave its use inside a reconstruction to future work. Our code, including dataset generation, training and evaluation, is available at \url{https://github.com/nmisyats/aurora-cnf}.

\section{Method}
\label{sec:method}

We learn a conditional distribution over auroral electron energy spectra,
$p(\mathbf{x} \mid \mathbf{c})$, where $\mathbf{x}\in\mathbb{R}^{n_E}$ is a log differential number flux discretized over $n_E$ fixed energy bins and $\mathbf{c}$ describes the geomagnetic conditions under which it was measured. The model is trained once on in-situ satellite data and is intended to be used as a prior in downstream inverse problem tasks.

\paragraph{Dataset}
\label{sec:data}

We use the science-quality DMSP SSJ/5 release of \citet{redmonNewDMSPDatabase2017a} obtained from NASA CDAWeb. Each record is an electron spectrum measured at \SI{1}{\hertz} by a satellite in a Sun-synchronous polar orbit over $n_E = 19$ logarithmically spaced channels with center energies covering \SI{30}{\electronvolt} to \SI{30}{\kilo\electronvolt}. Contrary to the raw instrument archive, this release supplies time-dependent geometric factors $G_k(t)$ that track detector degradation.

The differential number flux in channel $k$, $j_k$ [\si{\per\centi\metre\squared\per\second\per\steradian\per\electronvolt}], is recovered from raw electron counts $N_k$ and the supplied radiation background $B_k$ as $j_k = (N_k - B_k)/G_k$. Channels with no signal above background are set to the counting floor, which is the flux of a single count $1/G_k$. We aim to model the distribution of the log-flux
\begin{equation}
  \mathbf{x} \;=\; \log_{10}\mathbf{j} \;=\; [\log_{10} j_1, \dots, \log_{10} j_{n_E}]^\top.
  \label{eq:x}
\end{equation}

Our dataset consists of 240 satellite-days from F16, F17 and F18 between 2003 and 2014,
chosen to spread the coverage over geomagnetic activity. We keep only the records in
the auroral oval of either hemisphere, i.e. $60^\circ < |\Lambda| < 80^\circ$. We hold out 10\% of the days of each satellite, which gives 216 training days and 24 held-out days (12 from F16, and 6 each from F17 and F18). Because consecutive 1 Hz records are strongly correlated, we assign each satellite-day entirely to one of the two sets.
Since the days were selected per satellite, 5 of the 24 held-out dates also occur in the
training set for another satellite. Retraining without the 16 training days within one day
of a held-out date changes the held-out negative log-likelihood (Section 3) of the records without floored channels by $-0.02$ to $+0.06$ nats over three seeds,
which is of the order of the seed spread.
Overall, this yields 5.23M training and 581K held-out spectra.

\paragraph{Conditioning variables}

We define the context as the four-dimensional vector
\begin{equation}
  \mathbf{c} = \left[\, \Lambda/90^\circ,\; \sin(2\pi\,\tau/24),\; \cos(2\pi\,\tau/24),\; K_p/9 \,\right]^\top,
\end{equation}
where $\Lambda$ is the altitude-adjusted corrected geomagnetic (AACGM) latitude and
$\tau$ is the magnetic local time (MLT), both provided at 1 s resolution. A sinusoidal
encoding is applied to $\tau$ because it is cyclic. Since position and time alone do not
allow the model to distinguish quiet from intense geomagnetic conditions, we also
append the planetary $K_p$ index, obtained from the hourly OMNI2 compilation \citep{kingSolarWindSpatial2005}.
Unlike $\Lambda$ and $\tau$, $K_p$ varies only in time.

\paragraph{Model}

We use a conditional RealNVP \citep{dinhDensityEstimationUsing2017} which consists in an invertible mapping
$f_\theta(\cdot\,;\mathbf{c}): \mathbb{R}^{19}\!\to\!\mathbb{R}^{19}$ with learnable parameters $\theta$, which maps the spectra $\mathbf{x}$ to a standard normal distribution. The log-likelihood is then given by
\begin{equation}
  \log p_\theta(\mathbf{x}\mid\mathbf{c})
  = \log \mathcal{N}\!\left(f_\theta(\mathbf{x};\mathbf{c});\,\mathbf{0},\mathbf{I}\right)
  + \log\left|\det \frac{\partial f_\theta}{\partial \mathbf{x}}\right| .
  \label{eq:logprob}
\end{equation}
We chose a flow over a VAE or a network emitting a mean and variance because the invertible mapping $f_\theta$ has zero representation error and thus spectra that differ greatly from the training set remain reachable
\citep{asimInvertibleGenerativeModels2025}.

We define $f_\theta$ as a sequence of 8 blocks each consisting of an ActNorm layer
\citep{kingmaGlowGenerativeFlow2018} followed by a conditional affine coupling whose
log-scale and shift are predicted by a multi-layer perceptron (MLP) with two hidden layers of width 128. Masks alternate between even and odd
indices. The predicted log-scale is bounded by $3\tanh(\cdot)$, and the last layer of each coupling MLP is initialized to zero so that every
block starts as the identity. The entire model contains $1.96\times10^{5}$ parameters.

\paragraph{Training}
\label{sec:training}

The energy spectra are standardized per channel with the statistics of the training set. In that set, 41\% of the channel values have no signal above background
(from 20\% at 30 eV to 81\% at 30 keV), and 46\% are equal to $1/G_k$ once the channels with
exactly one count above background are included. These values form point masses that cannot be represented by
a continuous density. We therefore dequantize \citep{theisNoteEvaluationGenerative2016} the spectra by
adding a Gaussian jitter of standard deviation $\sigma = 0.05$ (in standardized units) to
the training spectra only, and quantify its effect in Section 3.

We train for 20 epochs with the Adam optimizer \citep{kingmaAdamMethodStochastic2017} and a batch size of 4096. The
learning rate is scheduled with linear warm-up over the first 300 steps followed by a cosine
decay, and the gradient norm is clipped to 5. We selected $4\times10^{-3}$ for the initial learning rate among $\{2, 4, 8, 16, 32\}\times10^{-3}$ after training 20 epochs on 195 training days and scoring on the 21 remaining ones. The training time of one flow is about two minutes on an RTX 4090 GPU.

\section{Results and Discussion}
\label{sec:results}

\paragraph{Evaluation protocol and baselines}
\label{sec:eval}

We evaluate the flow on the 24 held-out satellite-days using the negative log-likelihood
(NLL) in nats per spectrum as a score, computed on the standardized $\log_{10} j$.
We compare the flow against three baselines fitted on the training days. The first two are
Gaussian models with full covariance, one without context and one with a mean $\mu$ linear in
$\mathbf{c}$. The third is the spectral smoothness penalty $C_{\mathrm{sm}}$ of
\citep{misyatsFastScalableReconstruction2026}, which our learned prior is intended to
replace.

$C_{\mathrm{sm}}$ penalizes the second derivative of $\log j$ with respect to $\log E$, which over discrete channels can be
written as $C_{\mathrm{sm}}(\mathbf{x}) = \mathbf{x}^{\top}\mathbf{A}\,\mathbf{x}$, with $\mathbf{A}$ fixed. Using \( C_{\mathrm{sm}}\) as a regularization term with strength $\lambda$ acts as a Gaussian prior with zero mean and precision matrix $2\lambda\mathbf{A}$, allowing us to compare it to the other three priors tested.
By construction, $C_{\mathrm{sm}}$ constrains the spectral curvature alone and is unchanged by a uniform shift of $\mathbf{x}$ and by a
linear tilt in $\log E$, making it indifferent to the overall flux level and to the spectral slope. We fit $\lambda$ by
maximum likelihood and give these two directions the empirical Gaussian distribution of the training
data, which favors this prior.

\paragraph{Held-out likelihood}

The jitter turns each point mass into a narrow peak of width close to $\sigma$ that can be learned by our flow. A held-out channel on the counting floor lies on this peak, and the NLL of the
flow decreases by about 2.0 nats per floored channel, while the baselines have no such peak.
The NLL over all held-out records therefore mostly reflects the jitter width
(Table~\ref{tab:jitter}), and we compare the models on the $16.5\%$ of held-out records with
no channel on the floor. The baselines are refit on the floor-free training records, which
favors them, while the flow is trained once on all the records.

\begin{figure}[h]
  \begin{minipage}[c]{0.39\textwidth}
    \caption{Comparison between the held-out negative log-likelihoods of the baselines and of
    the flow with three contexts, on the floor-free test records. The
    baselines are fitted on floor-free training records, and the flows are trained on all
    training records. Flow values are means over three seeds, whose standard deviations are smaller than the markers (at
    most 0.04 nats).}
  \label{fig:nll}
  \end{minipage}\hfill
  \begin{minipage}[c]{0.6\textwidth}
    \includegraphics[width=\textwidth]{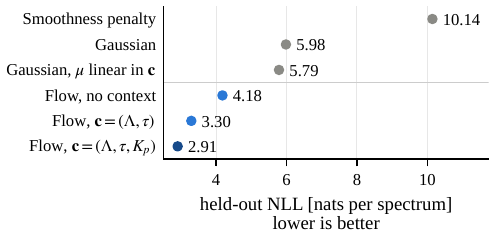}
  \end{minipage}
\end{figure}

Figure~\ref{fig:nll} shows that the flow assigns these spectra a higher density than all
baselines, with an NLL of 2.91 nats compared to 5.79 for the best Gaussian model, and 10.14 for the
smoothness penalty. The flow thus rates a spectrum 18 times more likely than the best
Gaussian and 1400 times more likely than the smoothness penalty. Without context, the flow already outperforms both
Gaussians (4.18 nats). Adding latitude and MLT lowers its NLL to 3.30 nats, while $K_p$ lowers it further to 2.91 nats.

\paragraph{Deformations}

Table~\ref{tab:deform} shows that the smoothness penalty gives nearly identical scores to a
real spectrum, a tilted spectrum, and an energy-reversed spectrum, while scoring a constant
spectrum as optimal. The flow instead assigns the lowest NLL to the real spectra and
penalizes all four deformations. The highly implausible energy-reversed spectrum shows a particularly high NLL.

\begin{table}[h]
  \begin{minipage}[c]{0.49\textwidth}
    \caption{Comparison between the smoothness penalty $C_{\mathrm{sm}}$ and the flow
    NLL on 4096 floor-free held-out spectra and four deformations of each spectrum.
    Values are medians over the spectra, and the tilts are applied to $\log_{10} j$ per
    decade of energy.}
    \label{tab:deform}
  \end{minipage}\hfill
  \begin{minipage}[c]{0.52\textwidth}
    \centering
    \begin{tabular}{lrr}
      \toprule
      candidate spectrum & $C_{\mathrm{sm}}$ & flow NLL \\
      \midrule
      real held-out spectra & $3.558$ & $2.3$ \\
      tilted by $+0.5$ & $3.558$ & $8.0$ \\
      tilted by $-0.5$ & $3.558$ & $16.5$ \\
      constant in $E$ & $0.000$ & $10.2$ \\
      reversed in energy & $3.557$ & $1.4\times10^{6}$ \\
      \bottomrule
    \end{tabular}
  \end{minipage}
\end{table}

\paragraph{Samples and conditioning}

\begin{figure}[h]
  \centering
  \includegraphics[width=0.49\textwidth]{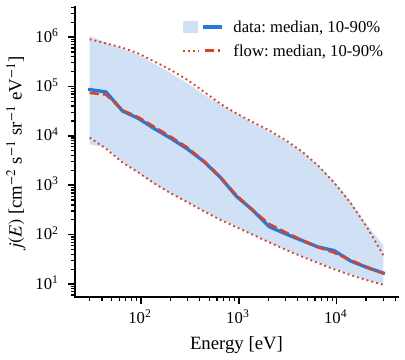}\hfill
  \includegraphics[width=0.49\textwidth]{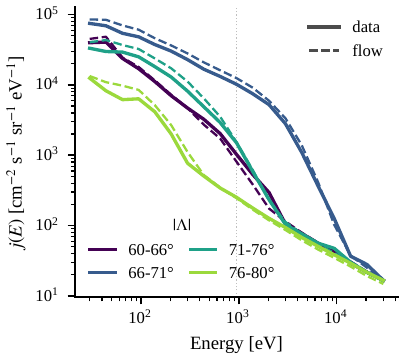}
  \caption{Visualization of the flow samples and of the held-out data. The median and the 10
  to $90\%$ range of $j(E)$ are shown for 20\,000 held-out records and for one flow sample at
  the context of each record (left). The median $j(E)$ of the held-out records (solid) and
  of one flow sample per record (dashed) is shown in four $|\Lambda|$ bands, in the southern
  oval between 0 and \SI{3}{\hour} MLT (right). The dotted line marks
  \SI{949}{\electronvolt}.}
  \label{fig:samples}
\end{figure}

Figure~\ref{fig:samples} (left) shows that we accurately recover the distribution of $j(E)$
at held-out contexts. The median and the 10 to $90\%$ range of the samples differ from the
data by at most 0.14 in $\log_{10} j$, except for the 90th percentile at
\SI{30}{\kilo\electronvolt}, where the difference reaches 0.20. Only one draw out of 20\,000
leaves the observed range of a channel by more than one decade. The jitter however places
$35\%$ of the draws slightly below the counting floor in at least one channel, by 0.017 in
$\log_{10} j$ in median.

Figure~\ref{fig:samples} (right) shows that we also recover the latitude structure of the
auroral oval on days the model has never seen. We group the held-out records of the
southern oval between 0 and \SI{3}{\hour} MLT into four bands of magnetic latitude,
with $|\Lambda|$ from $60^\circ$ to $66^\circ$ (equatorward), $66^\circ$ to $71^\circ$, $71^\circ$ to
$76^\circ$ and $76^\circ$ to $80^\circ$ (poleward).
The median flux in the \SI{949}{\electronvolt} channel rises by one order of magnitude from the equatorward band into the oval core and falls again poleward.
In the three bands poleward of $66^\circ$, the median of the samples is within 0.10 of the
data in $\log_{10} j$. In the equatorward band, it is 0.10 to 0.45 lower than the data,
depending on the seed. In the most poleward band, $73\%$ of the \SI{949}{\electronvolt} values sit on the counting
floor, and the median of both the data and the samples is therefore the floor value itself.

\paragraph{Effect of the jitter}

Table~\ref{tab:jitter} shows the effect of the jitter width $\sigma$.
Without jitter, the flow concentrates its density on the floor values, and the NLL over all held-out records decreases down to $-47.5$ nats after 20 epochs. On the floor-free records, the NLL however degrades to 3.93 nats, compared to 2.91 nats with $\sigma = 0.05$.
With jitter, the NLL over all records rises
from $-23.3$ to $0.8$ nats as $\sigma$ increases from 0.01 to 0.2, while the floor-free NLL
stays between 2.75 and 2.91 nats for $\sigma \le 0.05$. The flow outperforms all baselines up
to $\sigma = 0.1$, and falls behind both Gaussians at $\sigma = 0.2$, where the jitter may also blur the spectral shape.

\begin{table}[t]
  \centering
  \caption{Comparison between the held-out NLL of flows trained with a Gaussian jitter of
  standard deviation $\sigma$ on all held-out records and on the
  floor-free records. The $\sigma = 0$ column corresponds to a single flow trained over 20 epochs. The other columns are averages over three initialization seeds.}
  \label{tab:jitter}
  \begin{tabular}{lrrrrrr}
    \toprule
    jitter $\sigma$ & $0$ & $0.01$ & $0.02$ & $0.05$ & $0.1$ & $0.2$ \\
    \midrule
    all held-out records & $-47.5$ & $-23.3$ & $-18.5$ & $-11.5$ & $-5.7$ & $0.8$ \\
    floor-free held-out records & $3.93$ & $2.81$ & $2.75$ & $2.91$ & $3.92$ & $6.70$ \\
    \bottomrule
  \end{tabular}
\end{table}

\paragraph{Limitations}

The prior is evaluated as a density model only, and its use inside a reconstruction in place
of the smoothness penalty remains to be tested. Furthermore, $46\%$ of the training
channel values are equal to the floor value and the jitter only smooths these point masses.
A Poisson observation model on the raw counts would instead remove them. Finally, all three
satellites are Sun-synchronous. Northern oval records span MLT \SIrange{4.1}{20.6}{\hour}
and the night sector is sampled only in the south. Our conditioning is therefore limited to these observation conditions.

\section{Conclusion}
We proposed a density model of auroral electron energy spectra using a conditional normalizing flow trained on 216 satellite-days of DMSP SSJ measurements and conditioned on magnetic latitude, magnetic local time and the $K_p$ index. On held-out satellite-days, the flow assigns the measured spectra with signal in every channel 18 times the density of the best Gaussian baseline and 1400 times that of the smoothness penalty. Its samples reproduce the observed flux distribution and the latitude structure of the auroral oval. Using the flow as a learned prior in place of a hand-tuned regularization in tomographic reconstructions is a work in progress.

\section*{Acknowledgments}

This research was supported by ``Strategic Research Projects'' grant from ROIS (Research Organization of Information and Systems).

\bibliographystyle{plainnat} % We choose the "plain" reference style
\bibliography{refs} % Entries are in the refs.bib file

\end{document}